\pdfoutput=1
\documentclass[final,5p,times,twocolumn]{elsarticle}
\usepackage[T1]{fontenc}
\usepackage[utf8]{inputenc}
\usepackage{amsmath,mathtools}
\usepackage{newtxtext,newtxmath}
\usepackage{graphicx,booktabs,multirow,array}
\usepackage{xcolor,microtype,subcaption,dblfloatfix,placeins,enumitem}
\usepackage{tikz}
\usetikzlibrary{arrows.meta,positioning,calc,fit,backgrounds}
\usepackage{hyperref}
\hypersetup{colorlinks=true,linkcolor=black,citecolor=black,urlcolor=black,pdfauthor={Qiancheng Zhou, Xiaodong Yue, Wenhua Zhang}}
\biboptions{authoryear,round}
\setcitestyle{citesep={,}}
\definecolor{copper}{HTML}{B8794F}
\definecolor{oablred}{HTML}{B24B48}
\definecolor{methodgray}{HTML}{87939B}
\definecolor{blue}{HTML}{477D9B}
\definecolor{slate}{HTML}{354652}
\definecolor{warmbg}{HTML}{FBF4EF}
\definecolor{coolbg}{HTML}{F0F5F8}
\definecolor{teal}{HTML}{2A8D85}
\definecolor{fny}{HTML}{C65D64}
\definecolor{fpy}{HTML}{D4AA4A}
\setlist[enumerate]{leftmargin=*,itemsep=3pt,topsep=4pt,parsep=0pt}
\newcommand{\sg}{\operatorname{sg}}
\newcommand{\ind}{\mathbb{1}}
\newcommand{\swatch}[2]{\tikz[baseline=-.5ex]{\draw[#1,line width=1.1pt,#2] (0,0)--(.42,0);}}
\definecolor{ink}{HTML}{142334}
\definecolor{warmfield}{HTML}{FFE5D4}
\definecolor{coolfield}{HTML}{E1F1FE}
\definecolor{warmtitle}{HTML}{BD501E}
\definecolor{cooltitle}{HTML}{174D8D}
\definecolor{orangegray}{HTML}{D57B3D}
\definecolor{pointred}{HTML}{EA4235}
\definecolor{warmorange}{HTML}{EF742C}
\definecolor{coolLayerOne}{HTML}{76AEC8}
\definecolor{coolLayerTwo}{HTML}{90B6C5}
\definecolor{coolLayerThree}{HTML}{A5A7BC}
\definecolor{warmLayerFront}{HTML}{DCA998}
\definecolor{coolLayerTop}{HTML}{CFDFE7}
\definecolor{coolLayerSide}{HTML}{7E9DAC}
\definecolor{warmLayerTop}{HTML}{F2D8C6}
\definecolor{warmLayerSide}{HTML}{B9856B}
\definecolor{figCheckedInk}{HTML}{26313B}
\definecolor{figCheckedRedtitle}{HTML}{8C5D3F}
\definecolor{figCheckedBluetitle}{HTML}{366A88}
\definecolor{figCheckedGreentitle}{HTML}{3F7056}
\definecolor{figCheckedPointred}{HTML}{B24B48}
\definecolor{figCheckedOrange}{HTML}{B8794F}
\definecolor{figCheckedGreyarrow}{HTML}{858E93}
\definecolor{figCheckedPinkborder}{HTML}{DEC8C5}
\definecolor{figCheckedPinkpanel}{HTML}{FCF8F7}
\definecolor{figCheckedPinkhint}{HTML}{F7EDEB}
\definecolor{figCheckedPinkbase}{HTML}{FCF8F7}
\definecolor{figCheckedOrangeborder}{HTML}{DCC5B5}
\definecolor{figCheckedOrangebase}{HTML}{FCF8F3}
\definecolor{figCheckedBlueborder}{HTML}{C2D1DB}
\definecolor{figCheckedBluebase}{HTML}{F2F7FA}
\definecolor{figCheckedGreenborder}{HTML}{C6D8CB}
\definecolor{figCheckedGreenbase}{HTML}{F3F9F4}
\definecolor{figCheckedGreenmint}{HTML}{F3F9F4}
\definecolor{figCheckedCloudblue}{HTML}{DDEAF9}
\definecolor{figCheckedWarmred}{HTML}{F1432B}
\definecolor{figCheckedWarmorange}{HTML}{B8794F}
\definecolor{figCheckedCoolbluehalo}{HTML}{DFE7F7}
\definecolor{figCheckedCoolOne}{HTML}{76ABC6}
\definecolor{figCheckedCoolTwo}{HTML}{94BCD0}
\definecolor{figCheckedCoolThree}{HTML}{AA9DC7}
\definecolor{figCheckedWarmOne}{HTML}{E6A489}
\definecolor{figCheckedCubetop}{HTML}{D3E8EF}
\definecolor{figCheckedCubeside}{HTML}{6386A1}
\definecolor{figCheckedWarmtop}{HTML}{F4D8C9}
\definecolor{figCheckedWarmside}{HTML}{C78268}
\definecolor{figCheckedJring0}{HTML}{FDEBE6}
\definecolor{figCheckedJring1}{HTML}{FBE5DD}
\definecolor{figCheckedJring2}{HTML}{FAD7CC}
\definecolor{figCheckedJring3}{HTML}{F8C8B9}
\definecolor{figCheckedJring4}{HTML}{F9B4A0}
\definecolor{figCheckedJring5}{HTML}{F9A387}
\definecolor{figCheckedJring6}{HTML}{FC8765}
\definecolor{figCheckedJring7}{HTML}{FB653F}
\definecolor{figCheckedJring8}{HTML}{FE3B1F}
\definecolor{figCheckedJhalo0}{HTML}{E6F1FD}
\definecolor{figCheckedJhalo1}{HTML}{DDEBFB}
\definecolor{figCheckedJhalo2}{HTML}{CFE3FB}
\definecolor{figCheckedBring0}{HTML}{D0E6FD}
\definecolor{figCheckedBring1}{HTML}{B4D7FC}
\definecolor{figCheckedBring2}{HTML}{8FC4F7}
\definecolor{figCheckedBring3}{HTML}{6AB1F1}
\definecolor{figCheckedBring4}{HTML}{3593E4}

\begin{document}
\raggedbottom
\begin{frontmatter}
\title{From Points to Targets: Online Foreground Affinity and Background Learning for Infrared Small Target Detection}
\author{Qiancheng Zhou}\author{Xiaodong Yue}\author{Wenhua Zhang\corref{cor1}}
\cortext[cor1]{Corresponding author.}
\begin{abstract}
Infrared small-target detection must identify a few faint target pixels in extensive background clutter while keeping false alarms low. Under single-point supervision, each target has only one labeled foreground pixel with unknown spatial extent during training, whereas the detector must recover complete targets from the infrared image without any point guidance at inference. Existing methods typically reconstruct regional pseudo-labels from local image cues or intermediate predictions, which can introduce boundary noise around blurry target edges. This raises a central question: can a detector learn target structure directly from sparse points without first turning uncertain boundaries into dense training targets? To answer this question, we propose Online Foreground Affinity and Background Learning (OABL), a single-stage framework tailored for point-supervised infrared small-target detection. First, for online foreground affinity learning, each annotated target point serves as a feature anchor, and local spatial proximity and intensity similarity weight the feature alignment between the point and surrounding pixels. This guides the network to learn target-related representations without assigning hard foreground labels to uncertain boundary pixels. Second, for online background learning, we exclude the ambiguous neighborhood around each target point from negative sampling and select high-response false alarms outside this region as active negative candidates. Averaging the background loss over this active set keeps the remaining target-like clutter influential as easy background pixels are suppressed during training. Jointly optimizing these objectives in a single stage updates the detector directly without generating or refreshing persistent pseudo-masks. Across three public datasets and the combined SIRST3 benchmark, OABL improves detection probability by 1.06--7.82 percentage points over matched PAL training at a false-alarm area rate of $2\times10^{-5}$ and avoids PAL's multi-stage full-pool label updates.
\end{abstract}


\end{frontmatter}

\section{Introduction}\label{sec:intro}

Infrared small-target detection (IRSTD) supports long-range surveillance, maritime search, and early warning, where missed targets and background false alarms directly affect detection reliability. True targets usually occupy only a few pixels. Their compact responses often fade into the surrounding background, while bright edges and local reflections can resemble targets. Reliable detection must therefore recover faint targets and reject target-like clutter over a much larger image area. Detectors based on local contrast and contextual features typically learn from dense masks \citep{dai2021,li2023dnanet,wu2023,zhao2025msda}, yet pixel-level masks are laborious to annotate around ambiguous boundaries. During point-supervised training, each target has only one labeled foreground pixel with unknown spatial extent, whereas at inference the detector must recover complete targets from the infrared image without any point guidance.

Existing point-supervised IRSTD methods commonly recover the missing target extent through intermediate regional supervision. Image-guided methods such as MCLC and MCGC estimate candidate regions from local structure \citep{li2023mclc,kou2024}, while LESPS and PAL refine labels with detector predictions across progressive training stages \citep{ying2023,yu2025}. Recent work further incorporates diffusion priors and sample rebalancing \citep{liu2026}. Despite their progress, learning a pixel-level detector from sparse points still faces two key challenges. The first challenge is learning complete target structure under ambiguous boundaries. Existing methods convert uncertain local evidence into explicit regional training targets before supervising the detector. Because faint infrared targets decay smoothly into the background, premature boundary decisions easily truncate weak target edges or absorb adjacent noise into the foreground mask. The second challenge is suppressing long-tailed target-like clutter across a vast background. Unlabeled pixels near an annotated point may belong to the target and cannot be naively treated as negatives. Meanwhile, outside the target neighborhood, abundant easy background pixels quickly dominate standard loss averaging and overwhelm the gradients of the small population of bright, target-like clutter responses.

Unlike existing approaches that largely adapt generic pseudo-labeling or region-growing methods from natural images to infrared small-target detection, we design our framework around the intrinsic foreground and background cues of infrared targets. Compact infrared targets exhibit strong local spatial concentration and internal feature similarity even when their visible boundaries remain indistinct. Rather than forcing these cues into fixed pixel-level pseudo-masks, we use spatial proximity and intensity similarity around each point anchor to weight how strongly point-to-pixel features are aligned, while the detector head learns the final foreground response. At the same time, to handle severe foreground--background imbalance and sparse high-brightness clutter, we use the annotated point location to exclude uncertain target neighborhoods and focus negative supervision on the remaining high-response clutter.

We develop this idea into Online Foreground Affinity and Background Learning (OABL), which couples online foreground affinity learning with active background suppression in a single training stage. During each forward pass, the detector's shared features provide point-to-pixel affinity, while local spatial and intensity support weights the alignment objective. The output head learns foreground probabilities from the resulting shared representation. On the background side, a spatial exclusion radius around each point protects neighborhoods that may contain unlabeled target pixels. Current predictions then identify the remaining high-response candidates outside this radius, and averaging the background loss over this active set preserves effective gradients on hard false alarms as easy background pixels disappear from training. All three objectives update the same detector parameters at every step.

We evaluate OABL on NUDT-SIRST, NUAA-SIRST, IRSTD-1K, and their combined SIRST3 benchmark. OABL achieves reliable target detection across strict false-alarm operating points through direct single-stage optimization of the original detector. Detailed analyses of learned feature relations and background responses explain how the two mechanisms shape the trained detector. Our main contributions in this work are summarized as follows:

\begin{enumerate}

\item \textbf{Direct single-stage learning from point annotations.} We propose OABL to train a pixel-level detector directly from one foreground point per target in a single stage. Integrating positive point supervision with online feature alignment and active background learning preserves the original inference architecture without intermediate mask maintenance.

\item \textbf{Prior-guided online foreground affinity learning.} Exploiting the compact radiation and local feature similarity of infrared targets, we introduce a point-anchored affinity objective in which spatial and intensity cues determine the weight of feature alignment. This shapes target features in the shared representation while the detector learns the boundary.

\item \textbf{Active background learning for residual clutter.} To address long-tailed background clutter in infrared scenes, we combine point-distance spatial exclusion with active-set loss normalization. The resulting objective prevents easy background from dominating optimization and keeps supervision focused on sparse target-like false alarms.

\end{enumerate}

\section{Related work}

Learning infrared targets from point annotations involves three closely related problems: obtaining useful regional supervision from sparse labels, learning feature relationships beyond the annotated locations, and handling abundant background responses during detector training. We review developments in point-supervised IRSTD, affinity-based learning, and background sample selection in this order.

\subsection{Point-supervised infrared small-target detection}

Deep infrared detectors improve the recognition of faint targets by combining local contrast with larger spatial context. Asymmetric Contextual Modulation (ACM) models interactions between local features and context \citep{dai2021}. DNANet and UIU-Net aggregate representations at multiple scales \citep{li2023dnanet,wu2023}, and MSDA-Net adds directional features \citep{zhao2025msda}. These networks are commonly trained with full target masks. Point supervision retains the pixel-wise prediction task while replacing dense training labels with sparse foreground coordinates. For compact infrared targets, the missing region must be learned from weak local evidence and the detector's evolving representation.

A major line of point-supervised IRSTD recovers intermediate regional labels from point annotations. MCLC and MCGC estimate target support from image structure through clustering and region growth \citep{li2023mclc,kou2024}. LESPS evolves labels using detector predictions, while PAL combines label refinement with progressive training \citep{ying2023,yu2025}. Local contrast initialization, teacher guidance, and diffusion priors provide further ways to construct or refine the regional targets \citep{yang2024,li2026teacher,liu2026}. These methods make the unannotated target area trainable by first estimating regional target labels.

Other studies reconsider what the detector predicts. SPIRE models a probabilistic target response from local image cues \citep{ni2026}. Point-to-Mask couples mask estimation with center and radius prediction \citep{gao2026}. These designs show that sparse points can support several forms of target supervision. The distinction relevant here is the role assigned to local evidence: it may define a regional target, shape a probability response, or weight the feature relations learned by the detector.

\subsection{Feature relationships under sparse supervision}

Feature relationships carry sparse supervision beyond labeled pixels. The early point-supervised segmentation study of \citet{bearman2016} established the value of sparse foreground labels. AffinityNet propagates weak foreground seeds using learned pixel affinities, IRNet models displacement and boundary relationships, and AFA obtains affinities from attention \citep{ahn2018,ahn2019,ru2022}. \citet{tang2018} further showed how image-derived structural constraints can enter the network objective directly. These approaches provide different ways for limited annotations and local structure to influence dense predictions.

Point2Mask uses optimal transport to assign pixels to annotated points and derives dense supervisory assignments \citep{li2023point2mask}. LAFR uses local affinities to refine point-derived target labels in infrared detection \citep{peng2025}. These studies demonstrate the usefulness of point-to-pixel relations for recovering region supervision. In OABL, point-anchored relations are computed within the detector's shared representation and weighted by local image cues during single-stage training.

\subsection{Background selection and supervision}

Online Hard Example Mining selects high-loss examples, while focal loss reduces the contribution of well-classified examples \citep{shrivastava2016,lin2017}. Both address the large population of easy negatives in dense detection. Online pixel bootstrapping applies related selection to dense segmentation \citep{wu2016}. GHM reweights examples according to gradient density, while PISA prioritizes informative detection samples using classification and localization quality \citep{li2019ghm,cao2020}. These methods show that a negative sample's contribution depends on both its difficulty and the normalization of the selected population.

Sparse labels add a second question to hard-example selection: which unlabeled pixels should provide negative supervision? U2PL extracts negative information from uncertain predictions in semi-supervised segmentation \citep{wang2022}. In infrared detection, FEST mines difficult edges, PAL uses difficulty-aware training in its progressive procedure, and Diffuse to Detect combines regional supervision with sample rebalancing \citep{zhao2025fest,yu2025,liu2026}. These studies link negative learning to evolving supervision.

With only target locations labeled, a strong response near a point may come from an unannotated part of the target, while a response farther away may reveal target-like clutter. Background learning must therefore decide where negatives can be selected and how to keep the remaining hard responses influential as their number falls. Point geometry and prediction difficulty jointly guide negative supervision.

\section{Method}\label{sec:method}

\begin{figure*}[!t]
\centering
\resizebox{\textwidth}{!}{%
% Figure 1 is assembled here from independently editable image panels.
\begin{tikzpicture}[x=1cm,y=1cm,>=Latex,
 every node/.style={font=\fontsize{8.4}{10}\selectfont,text=figCheckedInk,inner sep=0pt,outer sep=0pt},
 tile/.style={inner sep=0,outer sep=0},
 head/.style={font=\fontsize{11.0}{12.7}\bfseries\selectfont},
 micro/.style={font=\fontsize{7.3}{8.7}\selectfont}]
\path[use as bounding box] (.12,.62) rectangle (18.60,7.64);

% Reference card frames. The large panels preserve the screenshot's exact bounds.
\fill[fill=figCheckedPinkbase,rounded corners=2.3mm]
  (.21,.71) rectangle (3.48,7.55);
\draw[rounded corners=2.3mm,draw=figCheckedPinkborder,line width=.47pt] (.21,.71) rectangle (3.48,7.55);
\fill[fill=figCheckedOrangebase,rounded corners=2.5mm]
  (6.61,4.39) rectangle (15.18,7.55);
\draw[rounded corners=2.5mm,draw=figCheckedOrangeborder,line width=.47pt] (6.61,4.39) rectangle (15.18,7.55);
\fill[fill=figCheckedBluebase,rounded corners=2.5mm]
  (6.61,.71) rectangle (15.18,4.21);
\draw[rounded corners=2.5mm,draw=figCheckedBlueborder,line width=.47pt] (6.61,.71) rectangle (15.18,4.21);
\fill[fill=figCheckedGreenmint,rounded corners=2.3mm]
  (15.35,.71) rectangle (18.50,7.55);
\draw[rounded corners=2.3mm,draw=figCheckedGreenborder,line width=.47pt] (15.35,.71) rectangle (18.50,7.55);

% Left problem statement
\node[anchor=west,font=\fontsize{9.5}{11.0}\bfseries\selectfont] at (.65,7.18) {Point annotation};

\node[tile] at (1.815,5.62) {\includegraphics[width=2.12cm,height=2.12cm,keepaspectratio]{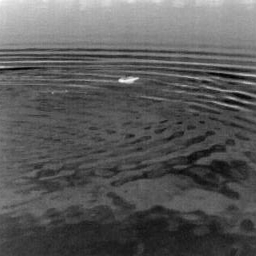}};
\draw[figCheckedOrange,dashed,line width=.64pt,dash pattern=on 2.5pt off 1.9pt]
  (1.574844,5.810469) rectangle (2.013750,6.249375);
\draw[figCheckedOrange,dashed,line width=.59pt,dash pattern=on 2.5pt off 1.9pt]
  (1.574844,5.810469)--(.975,4.40) (2.013750,5.810469)--(2.655,4.40);
\draw[figCheckedPointred,line width=1.05pt] (1.694297,6.029922)--(1.894297,6.029922) (1.794297,5.929922)--(1.794297,6.129922);
\node[tile] at (1.815,3.56) {\includegraphics[width=1.68cm,height=1.68cm,keepaspectratio]{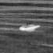}};
\draw[figCheckedPointred,line width=1.05pt] (1.715,3.56)--(1.915,3.56) (1.815,3.46)--(1.815,3.66);

\node[anchor=center,font=\fontsize{7.15}{8.6}\selectfont] at (1.815,2.54) {Annotated target};
\fill[fill=figCheckedPinkhint,rounded corners=1.7mm]
  (.40,.84) rectangle (3.27,2.37);
\foreach \yy/\word in {2.13/{Target extremely small},1.79/{Extent unknown},1.45/{Complex background},1.11/{High false alarms}} {
  \draw[figCheckedPointred,line width=1.15pt,line cap=round] (.545,\yy-.055)--(.655,\yy+.055) (.545,\yy+.055)--(.655,\yy-.055);
  \node[anchor=west,text=figCheckedRedtitle,font=\fontsize{7.2}{8.6}\selectfont] at (.83,\yy) {\word};
}

% Four-layer detector
\node[anchor=center,font=\fontsize{7.3}{8.2}\selectfont] at (4.31,5.40) {Shared};
\node[anchor=center,font=\fontsize{7.3}{8.2}\selectfont] at (4.31,5.11) {detector};
\node[tile] at (4.31,4.13) {\includegraphics[width=1.23cm,height=1.47cm,keepaspectratio]{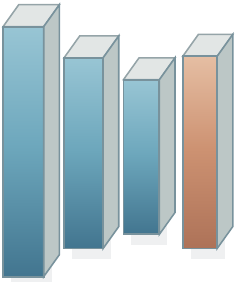}};

\node[anchor=center,micro] at (5.85,6.47) {Feature $F$};
\node[tile] at (5.85,5.45) {\includegraphics[width=1.40cm,height=1.40cm,keepaspectratio]{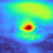}};
\node[anchor=center,micro] at (5.85,3.25) {Prediction $q$};
\node[tile] at (5.85,2.19) {\includegraphics[width=1.40cm,height=1.40cm,keepaspectratio]{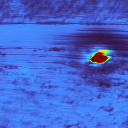}};

\fill[figCheckedGreyarrow]
 (3.48,4.075)--(3.565,4.075)--(3.565,4.025)--(3.70,4.13)--(3.565,4.235)--(3.565,4.185)--(3.48,4.185)--cycle;
\fill[figCheckedGreyarrow]
 (4.925,2.19)--(5.035,2.19)--(5.035,5.45)--(4.925,5.45)
 --(4.925,4.185)--(4.885,4.185)--(4.885,4.075)--(4.925,4.075)--cycle;
\foreach \yy in {5.45,2.19} {
 \fill[figCheckedGreyarrow]
  (4.98,\yy-.055)--(5.025,\yy-.055)--(5.025,\yy-.105)--(5.15,\yy)--(5.025,\yy+.105)--(5.025,\yy+.055)--(4.98,\yy+.055)--cycle;
 \fill[figCheckedGreyarrow]
  (6.55,\yy-.055)--(6.66,\yy-.055)--(6.66,\yy-.11)--(6.80,\yy)--(6.66,\yy+.11)--(6.66,\yy+.055)--(6.55,\yy+.055)--cycle;
}

% Foreground callout
\fill[fill=figCheckedWarmorange] (7.03,7.18) circle (.245);
\node[font=\fontsize{10}{11.5}\bfseries\selectfont,text=white] at (7.03,7.17) {1};
\node[anchor=west,font=\fontsize{9.5}{11.0}\bfseries\selectfont,text=figCheckedRedtitle] at (7.55,7.17)
  {Foreground: point-anchored affinity learning};
\node[anchor=center,micro] at (7.76,6.47) {Point affinity $A$};
\node[anchor=center,micro] at (9.65,6.47) {Local support $W$};
\node[anchor=center,font=\fontsize{7.3}{8.7}\selectfont] at (12.70,6.47) {Prior-weighted feature alignment};
\node[tile] at (7.76,5.45) {\includegraphics[width=1.40cm,height=1.40cm,keepaspectratio]{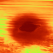}};
\node[tile] at (9.65,5.45) {\includegraphics[width=1.40cm,height=1.40cm,keepaspectratio]{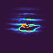}};

\foreach \xx in {7.76,9.65} {
  \draw[figCheckedPointred,line width=.85pt] (\xx-.07,5.45)--(\xx+.07,5.45)
  (\xx,5.38)--(\xx,5.52);
}

\node[tile] at (12.70,5.36) {\includegraphics[width=3.00cm,height=1.76cm,keepaspectratio]{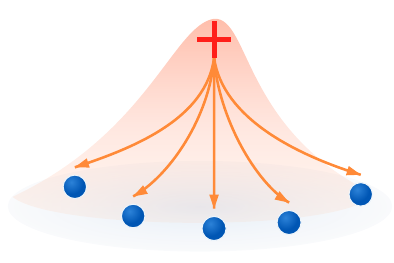}};

% Background callout
\fill[fill=figCheckedBluetitle] (7.03,3.81) circle (.245);
\node[text=white,font=\fontsize{10}{11.5}\bfseries\selectfont] at (7.03,3.81) {2};
\node[anchor=west,text=figCheckedBluetitle,font=\fontsize{9.0}{10.65}\bfseries\selectfont] at (7.58,3.80)
 {Background: selective suppression of residual clutter};
\node[anchor=center,micro] at (7.76,3.25) {Candidate set $E$};
\node[anchor=center,micro] at (9.65,3.25) {Prediction $q$};
\node[anchor=center,micro] at (11.54,3.25) {Active set $H$};
\node[tile] at (7.76,2.19) {\includegraphics[width=1.40cm,height=1.40cm,keepaspectratio]{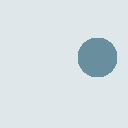}};
\node[tile] at (9.65,2.19) {\includegraphics[width=1.40cm,height=1.40cm,keepaspectratio]{figures/fig1_zoom_q_background.png}};
\node[tile] at (11.54,2.19) {\includegraphics[width=1.40cm,height=1.40cm,keepaspectratio]{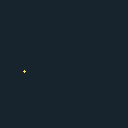}};

\node[tile] at (13.91,2.39) {\includegraphics[width=2.43cm,height=1.16cm,keepaspectratio]{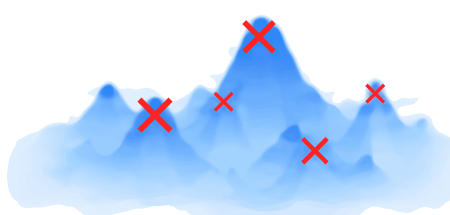}};

% Green panel: Joint optimization
\fill[fill=figCheckedGreentitle!80!black] (15.72,7.18) circle (.235);
\node[text=white,font=\fontsize{9.7}{11.5}\bfseries\selectfont] at (15.72,7.18) {3};
\node[anchor=west,text=figCheckedGreentitle,font=\fontsize{8.42}{10.1}\bfseries\selectfont]
  at (16.08,7.18) {Joint optimization};
\node[anchor=center,micro] at (16.94,6.72) {Single-stage training};

\begin{scope}[shift={(16.94,5.45)}]
\foreach \factor/\colorname/\opacity in {
  1.01/figCheckedJhalo0/.31,.91/figCheckedJhalo1/.25,.81/figCheckedJhalo2/.22,
  .83/figCheckedJring0/.28,.73/figCheckedJring1/.33,.64/figCheckedJring2/.35,
  .57/figCheckedJring3/.38,.50/figCheckedJring4/.47,.43/figCheckedJring5/.52,
  .36/figCheckedJring6/.58,.29/figCheckedJring7/.67,.22/figCheckedJring8/.82} {
  \begin{scope}[scale=\factor]
   \path[fill=\colorname,opacity=\opacity]
    (0,.93)..controls(.13,.92)and(.25,.49)..(.53,.28)
    ..controls(.81,.10)and(1.02,.07)..(1.03,0)
    ..controls(.99,-.10)and(.72,-.29)..(.48,-.49)
    ..controls(.20,-.72)and(.10,-.91)..(0,-.91)
    ..controls(-.10,-.91)and(-.20,-.72)..(-.48,-.49)
    ..controls(-.72,-.29)and(-.99,-.10)..(-1.03,0)
    ..controls(-1.02,.07)and(-.81,.10)..(-.53,.28)
    ..controls(-.25,.49)and(-.13,.92)..cycle;
  \end{scope}
}
\draw[white,line width=1.10pt] (0,-.128)--(0,.128) (-.128,0)--(.128,0);
\end{scope}

\node[anchor=center,align=center,text=figCheckedInk,font=\fontsize{8.2}{9.8}\selectfont]
  at (16.94,4.37) {One gradient update};

\node[anchor=center,micro] at (16.94,3.91) {Target prediction};
\node[tile] at (16.94,2.99) {\includegraphics[width=1.42cm,height=1.42cm,keepaspectratio]{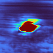}};
\fill[fill=white,rounded corners=1.7mm]
 (15.50,.89) rectangle (18.34,2.14);
\foreach \yy/\claim in {1.94/{Point-anchored affinity},1.59/{Active background},1.24/{Single-stage learning}} {
  \fill[fill=figCheckedGreentitle!87!black] (15.74,\yy) circle (.12);
  \draw[white,line width=.86pt] (15.69,\yy)--(15.735,\yy-.045)--(15.81,\yy+.055);
  \node[anchor=west,text=figCheckedGreentitle,font=\fontsize{7.2}{8.6}\selectfont] at (15.94,\yy) {\claim};
}
\end{tikzpicture}
}
\caption{Overview of the proposed OABL framework. Given single-point annotations, OABL trains a shared detector through three components: (1) Foreground point-anchored affinity learning, which uses local spatial and intensity support $W$ around each annotated point to weight online feature affinity $A$ and capture target structure, (2) Background selective suppression, which excludes uncertain target neighborhoods to form the candidate set $E$ and focuses supervision on high-response active clutter $H$, and (3) Joint optimization, which updates the shared detector end-to-end in a single training stage.}
\label{fig:framework}
\end{figure*}

\subsection{Problem formulation}

We consider point-supervised infrared small-target detection with one annotated foreground location for each target instance. All other pixels in that target remain unlabeled. Fig.~\ref{fig:framework} summarizes the OABL training framework. In a mini-batch of $B$ images, let $\mathcal P_b$ be the points in image $x_b$, $u=(i,j)$ a pixel on the prediction grid $\Omega$, and $p=(i_p,j_p)$ an annotated position. The detector $f_\theta$ produces shared features $F_b\in\mathbb R^{D\times H\times W}$ and foreground logits $z_b\in\mathbb R^{H\times W}$. The foreground probability is $q_b(u)=\sigma(z_b(u))$, where $\sigma$ denotes the sigmoid function. Features are interpolated to the prediction grid when needed, and all point coordinates and distances use this grid. At inference, the detector segments complete targets directly from the infrared image without any point guidance.

\begin{equation}\label{eq:point}
\mathcal L_{\mathrm{point}}=-\frac{\sum_{b=1}^{B}\sum_{p\in\mathcal P_b}\log q_b(p)}{\sum_{b=1}^{B}|\mathcal P_b|+\epsilon}.
\end{equation}

The positive point loss in Eq.~\eqref{eq:point} establishes the known target locations as foreground, where $\epsilon=10^{-6}$ ensures numerical stability. Because a single point leaves the target extent unknown, we next introduce online foreground affinity learning and active background learning around these points.

\subsection{Online foreground affinity learning}\label{sec:method-affinity}

\paragraph{Online point-to-pixel affinity} Every annotated target point provides a feature anchor from the current detector. We compare its normalized feature with those of other pixels using cosine similarity, and retain the maximum when several target points are present. The similarity margin $m$ and sharpening exponent $\gamma$ in Eq.~\eqref{eq:affinity} focus alignment on selected feature relationships. Because the anchors and compared features belong to the same trainable representation, a point can influence features beyond its labeled coordinate.

\begin{equation}\label{eq:affinity}\begin{aligned}
\overline F_b(u)&=\frac{F_b(u)}{\max(\|F_b(u)\|_2,\epsilon_F)},\\
s_b(u)&=\max_{p\in\mathcal P_b}\overline F_b(p)^\top\overline F_b(u),\\
A_b(u)&=\left(\frac{[s_b(u)-m]_+}{1-m}\right)^\gamma,\qquad [v]_+=\max(v,0).
\end{aligned}\end{equation}

\paragraph{Local support from infrared appearance} Compact infrared targets often show spatially concentrated responses and similar intensity within the target core. These cues can guide which point-to-pixel relations deserve stronger learning, even when the target boundary is uncertain. We combine a spatial Gaussian kernel centered on annotated points with an intensity Gaussian kernel comparing pixel values. Eq.~\eqref{eq:support} takes the maximum of each kernel over points and multiplies the two resulting maps. The support $W_b$ is fixed for the current image and weights feature alignment in the next step. Here $g_b(u)\in[0,1]$ denotes normalized grayscale intensity. The spatial and intensity maxima are computed independently over the point set.

\begin{equation}\label{eq:support}\begin{aligned}
S_b(u)&=\max_{p\in\mathcal P_b}\exp\!\left(-\frac{\|u-p\|_2^2}{2\sigma_s^2}\right),\\
I_b(u)&=\max_{p\in\mathcal P_b}\exp\!\left(-\frac{(g_b(u)-g_b(p))^2}{2\sigma_I^2}\right),\\
W_b(u)&=S_b(u)I_b(u).
\end{aligned}\end{equation}

\paragraph{Prior-weighted feature alignment} The image support $W_b(u)$ indicates how strongly a location should participate in feature alignment. In Eq.~\eqref{eq:affloss}, it multiplies the penalty that brings the learned affinity $A_b(u)$ toward one. The value of $W_b(u)$ therefore controls the strength of a relation update. For example, $W_b(u)=0.2$ gives that relation one fifth of the weight of a fully supported relation, while the final foreground probability $q_b(u)$ is learned by the shared detector under point supervision and background discrimination.

\begin{equation}\label{eq:affloss}
\mathcal L_{\mathrm{aff}}=\frac{\sum_{b,u}W_b(u)\bigl(1-A_b(u)\bigr)^2}{\sum_{b,u}W_b(u)+\epsilon}.
\end{equation}

For a point-to-pixel pair above the similarity margin $m$, minimizing Eq.~\eqref{eq:affloss} increases its feature similarity to the selected point anchor. As shown in Eq.~\eqref{eq:affgrad}, the gradient is weighted by $W_b(u)$, so pixels with stronger local image support contribute more to the update. The detector learns foreground scores from these shared features through the combined point and background objectives. Furthermore, the product $W_bA_b$ combines learned affinity with local image support, which Section~\ref{sec:affinity-analysis} compares alongside foreground predictions in models trained with and without $\mathcal L_{\mathrm{aff}}$ (Fig.~\ref{fig:affinity}).

\begin{equation}\label{eq:affgrad}
\frac{\partial\mathcal L_{\mathrm{aff}}}{\partial s_b(u)}=
-\frac{2\gamma W_b(u)}{Z_W(1-m)}\bigl(1-A_b(u)\bigr)r^{\gamma-1},
\end{equation}
where $Z_W=\sum_{b,u}W_b(u)+\epsilon$ and $r=(s_b(u)-m)/(1-m)$ for $m<s_b(u)<1$.

\subsection{Active background learning for residual clutter}\label{sec:method-bg}

A point marks one known foreground location, and nearby pixels may belong to the same target. We therefore preserve a neighborhood around each point for target learning. Outside these neighborhoods, the detector encounters a large pool of background candidates, including a small number of high-response structures that resemble true targets. As training suppresses easy background, the remaining high responses become the pixels most worth correcting. This motivates a spatially eligible, prediction-driven active background loss.

\paragraph{Spatial eligibility and active response selection} We use the distance to the closest annotated point to define eligible background pixels, $E_b=\{u:d_b(u)\ge R\}$. The radius $R$ excludes an uncertain region around each target point. Within $E_b$, an eligible pixel is active when its predicted foreground probability exceeds $\tau$. Eq.~\eqref{eq:active} gives the resulting mask $H_b$, with selection computed from detached predictions. The candidate region describes where negative supervision may act, and the current prediction identifies which responses still require correction.

\begin{equation}\label{eq:active}\begin{aligned}
d_b(u)&=\min_{p\in\mathcal P_b}\|u-p\|_2,\\
H_b(u)&=\ind[d_b(u)\ge R]\,\ind[\sg(q_b(u))>\tau].
\end{aligned}\end{equation}
Here $\ind[\cdot]$ is the indicator and $\sg(\cdot)$ stops gradients.

\paragraph{Active normalization for residual clutter} The background objective in Eq.~\eqref{eq:bgloss} minimizes the mean foreground probability of active pixels. As the detector suppresses easy negatives, fewer pixels remain in $H_b$. Averaging over all eligible background pixels $N_E=\sum_b |E_b|$ would scale down the contribution of those remaining responses by the shrinking active fraction $N_H/N_E$ and dilute gradients on residual false alarms. Averaging over the active count $N_H$ counteracts this shrinkage. The gradient in Eq.~\eqref{eq:bggrad} shows this selected-pixel weight explicitly so the loss stays focused on responses that remain difficult.

\begin{equation}\label{eq:bgloss}
\mathcal L_{\mathrm{bg}}=\frac{\sum_{b,u}H_b(u)q_b(u)}{N_H+\epsilon},\qquad N_H=\sum_{b,u}H_b(u).
\end{equation}

\begin{equation}\label{eq:bggrad}
\frac{\partial\mathcal L_{\mathrm{bg}}}{\partial z_b(u)}=
\frac{H_b(u)}{N_H+\epsilon}\,q_b(u)\bigl(1-q_b(u)\bigr).
\end{equation}

Distance protects the uncertain target neighborhood, while predicted activity selects hard background responses elsewhere. Both the affinity and background objectives update the shared detector, as illustrated in Fig.~\ref{fig:framework}.

\subsection{Unified single-stage optimization}

\begin{equation}\label{eq:objective}
\mathcal L=\lambda_p\mathcal L_{\mathrm{point}}+\lambda_a\mathcal L_{\mathrm{aff}}+\lambda_b\mathcal L_{\mathrm{bg}}.
\end{equation}

\begin{table*}[t]\centering\footnotesize
\caption{Quantitative comparison with state-of-the-art point-supervised methods across four IRSTD benchmarks ($P_d$ and mIoU in $\%$, $F_a$ in $10^{-5}$). OABL achieves superior detection performance under both the strict false-alarm budget ($F_a=2\times10^{-5}$) and the default threshold of 0.5. Values are means and standard deviations over three seeds, and bold indicates the best point-supervised result.}\label{tab:main}
\setlength{\tabcolsep}{6pt}\renewcommand{\arraystretch}{1.06}
\begin{tabular}{llrrrr}\toprule
Dataset & Method & $P_d$ at fixed $F_a\uparrow$ & $P_d\uparrow$ & $F_a\downarrow$ & mIoU$\uparrow$\\\midrule
SIRST3 & LESPS & $92.32\,\pm\,0.63$ & $94.11\,\pm\,0.37$ & $4.59\,\pm\,0.88$ & $53.76\,\pm\,2.06$\\
 & PAL & $96.22\,\pm\,1.03$ & $95.88\,\pm\,0.81$ & $1.73\,\pm\,0.30$ & $65.30\,\pm\,2.39$\\
 & MCLC & $92.39\,\pm\,3.57$ & $94.71\,\pm\,1.10$ & $3.27\,\pm\,0.36$ & $\mathbf{66.03\,\pm\,0.80}$\\
 & Diffuse & --- & $92.18\,\pm\,6.53$ & $3.79\,\pm\,2.35$ & $64.13\,\pm\,2.14$\\
 & \textbf{OABL} & $\mathbf{97.28\,\pm\,0.40}$ & $\mathbf{97.08\,\pm\,0.37}$ & $\mathbf{1.03\,\pm\,0.24}$ & $63.44\,\pm\,0.67$\\
 & Full mask & $97.97\,\pm\,0.23$ & $97.81\,\pm\,0.58$ & $2.15\,\pm\,0.88$ & $81.69\,\pm\,0.74$\\
\midrule
NUDT-SIRST & LESPS & $95.53\,\pm\,0.80$ & $96.47\,\pm\,0.40$ & $3.86\,\pm\,0.28$ & $55.86\,\pm\,1.67$\\
 & PAL & $97.34\,\pm\,1.21$ & $96.12\,\pm\,1.92$ & $1.25\,\pm\,0.30$ & $67.05\,\pm\,8.36$\\
 & MCLC & $93.81\,\pm\,1.50$ & $94.71\,\pm\,0.64$ & $3.40\,\pm\,1.66$ & $66.89\,\pm\,0.31$\\
 & Diffuse & --- & $95.10\,\pm\,1.86$ & $2.10\,\pm\,1.17$ & $\mathbf{69.00\,\pm\,1.47}$\\
 & \textbf{OABL} & $\mathbf{98.48\,\pm\,0.22}$ & $\mathbf{98.24\,\pm\,0.16}$ & $\mathbf{0.65\,\pm\,0.46}$ & $62.46\,\pm\,1.52$\\
 & Full mask & $98.98\,\pm\,0.32$ & $98.52\,\pm\,0.28$ & $0.28\,\pm\,0.08$ & $93.93\,\pm\,0.43$\\
\midrule
NUAA-SIRST & LESPS & $88.80\,\pm\,2.63$ & $91.89\,\pm\,0.22$ & $4.89\,\pm\,2.93$ & $57.81\,\pm\,1.63$\\
 & PAL & $86.60\,\pm\,4.69$ & $85.93\,\pm\,4.99$ & $1.89\,\pm\,0.48$ & $55.67\,\pm\,6.89$\\
 & MCLC & $91.88\,\pm\,3.10$ & $92.02\,\pm\,1.01$ & $2.30\,\pm\,1.10$ & $71.39\,\pm\,1.06$\\
 & Diffuse & $\mathbf{94.42\,\pm\,0.58}$ & $\mathbf{94.93\,\pm\,0.58}$ & $2.15\,\pm\,0.56$ & $\mathbf{71.48\,\pm\,1.55}$\\
 & \textbf{OABL} & $\mathbf{94.42\,\pm\,0.79}$ & $93.54\,\pm\,0.76$ & $\mathbf{1.85\,\pm\,0.46}$ & $56.47\,\pm\,2.16$\\
 & Full mask & $88.42\,\pm\,9.22$ & $94.68\,\pm\,2.66$ & $6.40\,\pm\,3.05$ & $70.12\,\pm\,6.39$\\
\midrule
IRSTD-1K & LESPS & $84.07\,\pm\,1.17$ & $88.22\,\pm\,1.47$ & $5.18\,\pm\,0.91$ & $47.03\,\pm\,3.31$\\
 & PAL & $87.31\,\pm\,3.34$ & $89.90\,\pm\,3.09$ & $3.01\,\pm\,1.44$ & $55.97\,\pm\,10.06$\\
 & MCLC & $89.22\,\pm\,1.39$ & $91.47\,\pm\,2.03$ & $3.09\,\pm\,1.17$ & $\mathbf{63.54\,\pm\,1.96}$\\
 & Diffuse & --- & $92.03\,\pm\,1.56$ & $\mathbf{1.47\,\pm\,1.25}$ & $61.23\,\pm\,0.92$\\
 & \textbf{OABL} & $\mathbf{93.38\,\pm\,1.52}$ & $\mathbf{93.15\,\pm\,1.40}$ & $2.12\,\pm\,0.71$ & $57.78\,\pm\,1.67$\\
 & Full mask & $90.28\,\pm\,3.48$ & $92.93\,\pm\,0.58$ & $3.11\,\pm\,1.41$ & $65.40\,\pm\,1.80$\\
\bottomrule\end{tabular}
\par\vspace{3pt}\begin{minipage}{.96\textwidth}\footnotesize
---: the evaluated threshold range does not bracket the budget. Full-mask supervision is included as an upper-bound reference.
\end{minipage}\end{table*}

OABL computes the positive point loss, online foreground affinity loss, and active background loss from the same forward pass. The three signals jointly update the network via Eq.~\eqref{eq:objective}. In the next pass, the current features provide renewed point relations and the current predictions select a new active background set while the annotations remain unchanged. For $K$ points on a $D$-channel grid of size $H\times W$, affinity comparisons cost $O(KDHW)$, spatial support costs $O(KHW)$, and background selection costs $O(HW)$ during training, with no parameters or overhead added at inference.

\section{Experiments}

We first ask whether direct single-stage point supervision can recover faint targets under strict false-alarm constraints. We then examine the two mechanisms that make this possible: how local priors shape target-related feature relations, and how active background normalization sustains supervision on residual clutter. Finally, training trajectories and runtime measurements examine single-stage optimization.

\subsection{Experimental setup}\label{sec:setup}

\paragraph{Datasets and point annotations} We use NUDT-SIRST, NUAA-SIRST, and IRSTD-1K, with 664, 214, and 201 test images \citep{li2023dnanet,dai2021,zhang2022}. SIRST3 combines the three datasets into a joint-training benchmark with 1,079 test images. Following the point protocol of \citet{ying2023,yu2025}, each connected target instance contributes one foreground point sampled inside its ground-truth mask. Dense masks are used to generate these points and to evaluate predictions. Detector training uses images and point labels, while inference uses the images without point guidance.

\begin{figure*}[!t]
\centering
{\footnotesize\swatch{blue}{dashed}\ PAL\quad\swatch{methodgray}{dash dot}\ MCLC\quad\swatch{oablred}{solid}\ OABL\par}\vspace{2pt}
\begin{subfigure}{.24\textwidth}\centering\includegraphics[width=\linewidth]{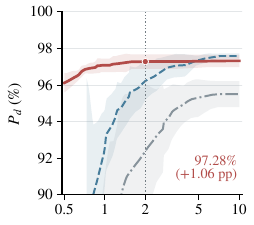}\caption{SIRST3 (joint training)}\end{subfigure}\hfill
\begin{subfigure}{.24\textwidth}\centering\includegraphics[width=\linewidth]{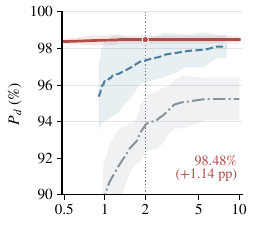}\caption{NUDT-SIRST}\end{subfigure}\hfill
\begin{subfigure}{.24\textwidth}\centering\includegraphics[width=\linewidth]{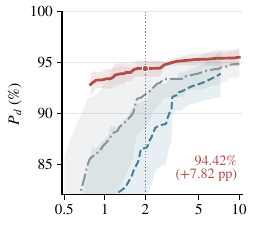}\caption{NUAA-SIRST}\end{subfigure}\hfill
\begin{subfigure}{.24\textwidth}\centering\includegraphics[width=\linewidth]{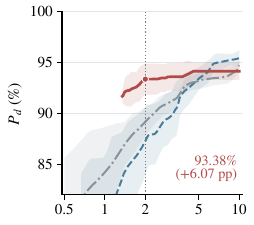}\caption{IRSTD-1K}\end{subfigure}
\par\vspace{2pt}{\footnotesize False-alarm area $F_a$ ($\times10^{-5}$)\par}
\caption{Detection probability ($P_d$) across false-alarm budgets ($F_a$) on four benchmarks. OABL maintains high detection probability over a broad range of low false-alarm rates and achieves substantial gains over PAL at the strict $F_a=2\times10^{-5}$ budget (vertical dashed line).}
\label{fig:detection}
\end{figure*}

\begin{figure*}[t]
\centering
{\footnotesize\textcolor{teal}{\rule{5pt}{5pt}}\ TP\qquad\textcolor{fny}{\rule{5pt}{5pt}}\ FN\qquad\textcolor{fpy}{\rule{5pt}{5pt}}\ FP\par}\vspace{3pt}
{\footnotesize\setlength{\tabcolsep}{3pt}
\begin{tabular}{@{}ccccc@{}}
Scene & Input + GT & PAL & MCLC & OABL\\[2pt]
\begin{minipage}[c][2.20cm][c]{2.20cm}\centering\includegraphics[width=2.20cm,height=2.20cm,keepaspectratio]{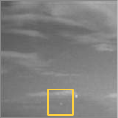}\end{minipage} & \begin{minipage}[c][2.20cm][c]{2.20cm}\centering\includegraphics[width=2.20cm,height=2.20cm,keepaspectratio]{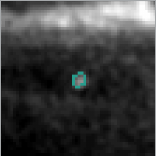}\end{minipage} & \begin{minipage}[c][2.20cm][c]{2.20cm}\centering\includegraphics[width=2.20cm,height=2.20cm,keepaspectratio]{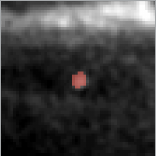}\end{minipage} & \begin{minipage}[c][2.20cm][c]{2.20cm}\centering\includegraphics[width=2.20cm,height=2.20cm,keepaspectratio]{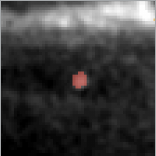}\end{minipage} & \begin{minipage}[c][2.20cm][c]{2.20cm}\centering\includegraphics[width=2.20cm,height=2.20cm,keepaspectratio]{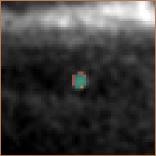}\end{minipage}\\
\multicolumn{5}{c}{(a) NUDT-SIRST: faint compact target}\\[4pt]
\begin{minipage}[c][2.20cm][c]{2.20cm}\centering\includegraphics[width=2.20cm,height=2.20cm,keepaspectratio]{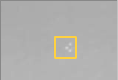}\end{minipage} & \begin{minipage}[c][2.20cm][c]{2.20cm}\centering\includegraphics[width=2.20cm,height=2.20cm,keepaspectratio]{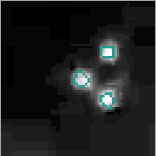}\end{minipage} & \begin{minipage}[c][2.20cm][c]{2.20cm}\centering\includegraphics[width=2.20cm,height=2.20cm,keepaspectratio]{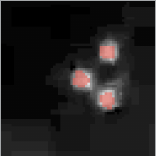}\end{minipage} & \begin{minipage}[c][2.20cm][c]{2.20cm}\centering\includegraphics[width=2.20cm,height=2.20cm,keepaspectratio]{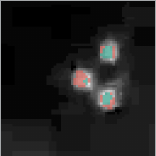}\end{minipage} & \begin{minipage}[c][2.20cm][c]{2.20cm}\centering\includegraphics[width=2.20cm,height=2.20cm,keepaspectratio]{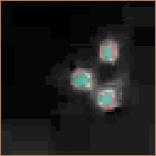}\end{minipage}\\
\multicolumn{5}{c}{(b) NUAA-SIRST: closely spaced targets}\\[4pt]
\begin{minipage}[c][2.20cm][c]{2.20cm}\centering\includegraphics[width=2.20cm,height=2.20cm,keepaspectratio]{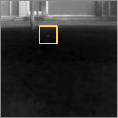}\end{minipage} & \begin{minipage}[c][2.20cm][c]{2.20cm}\centering\includegraphics[width=2.20cm,height=2.20cm,keepaspectratio]{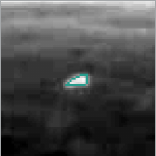}\end{minipage} & \begin{minipage}[c][2.20cm][c]{2.20cm}\centering\includegraphics[width=2.20cm,height=2.20cm,keepaspectratio]{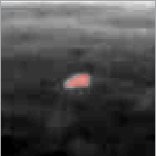}\end{minipage} & \begin{minipage}[c][2.20cm][c]{2.20cm}\centering\includegraphics[width=2.20cm,height=2.20cm,keepaspectratio]{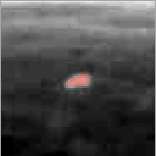}\end{minipage} & \begin{minipage}[c][2.20cm][c]{2.20cm}\centering\includegraphics[width=2.20cm,height=2.20cm,keepaspectratio]{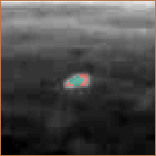}\end{minipage}\\
\multicolumn{5}{c}{(c) IRSTD-1K: elongated target}\\[4pt]
\end{tabular}}
\caption{Qualitative comparison on challenging infrared scenes. Compared with PAL and MCLC, OABL accurately recovers faint compact targets (a), separates closely spaced targets (b), and captures elongated target structures (c) while suppressing background false alarms. Green, red, and yellow indicate true-positive (TP), false-negative (FN), and false-positive (FP) pixels.}
\label{fig:qualitative}
\end{figure*}

\begin{figure*}[t]
\centering
{\footnotesize\setlength{\tabcolsep}{3pt}
\begin{tabular}{@{}ccccc@{}}
&\multicolumn{2}{c}{\textbf{Point-supported feature relations $WA$}}&\multicolumn{2}{c}{\textbf{Detector prediction $q$}}\\[3pt]
Input & Without affinity & With affinity & Without affinity & With affinity\\[2pt]
\begin{minipage}[c][2.20cm][c]{2.20cm}\centering\includegraphics[width=2.20cm,height=2.20cm,keepaspectratio]{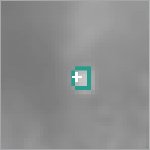}\end{minipage} & \begin{minipage}[c][2.20cm][c]{2.20cm}\centering\includegraphics[width=2.20cm,height=2.20cm,keepaspectratio]{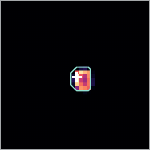}\end{minipage} & \begin{minipage}[c][2.20cm][c]{2.20cm}\centering\includegraphics[width=2.20cm,height=2.20cm,keepaspectratio]{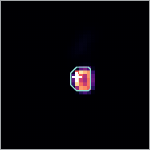}\end{minipage} & \begin{minipage}[c][2.20cm][c]{2.20cm}\centering\includegraphics[width=2.20cm,height=2.20cm,keepaspectratio]{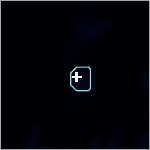}\end{minipage} & \begin{minipage}[c][2.20cm][c]{2.20cm}\centering\includegraphics[width=2.20cm,height=2.20cm,keepaspectratio]{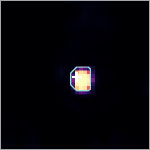}\end{minipage}\\[2pt]
\begin{minipage}[c][2.20cm][c]{2.20cm}\centering\includegraphics[width=2.20cm,height=2.20cm,keepaspectratio]{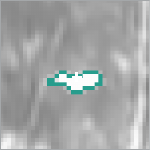}\end{minipage} & \begin{minipage}[c][2.20cm][c]{2.20cm}\centering\includegraphics[width=2.20cm,height=2.20cm,keepaspectratio]{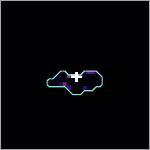}\end{minipage} & \begin{minipage}[c][2.20cm][c]{2.20cm}\centering\includegraphics[width=2.20cm,height=2.20cm,keepaspectratio]{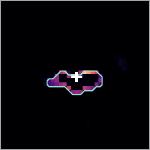}\end{minipage} & \begin{minipage}[c][2.20cm][c]{2.20cm}\centering\includegraphics[width=2.20cm,height=2.20cm,keepaspectratio]{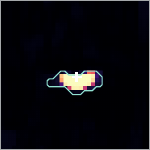}\end{minipage} & \begin{minipage}[c][2.20cm][c]{2.20cm}\centering\includegraphics[width=2.20cm,height=2.20cm,keepaspectratio]{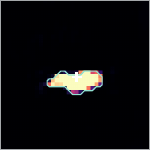}\end{minipage}\\[2pt]
\begin{minipage}[c][2.20cm][c]{2.20cm}\centering\includegraphics[width=2.20cm,height=2.20cm,keepaspectratio]{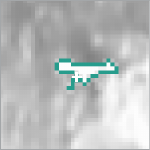}\end{minipage} & \begin{minipage}[c][2.20cm][c]{2.20cm}\centering\includegraphics[width=2.20cm,height=2.20cm,keepaspectratio]{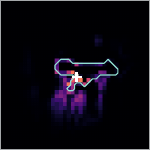}\end{minipage} & \begin{minipage}[c][2.20cm][c]{2.20cm}\centering\includegraphics[width=2.20cm,height=2.20cm,keepaspectratio]{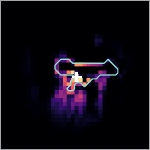}\end{minipage} & \begin{minipage}[c][2.20cm][c]{2.20cm}\centering\includegraphics[width=2.20cm,height=2.20cm,keepaspectratio]{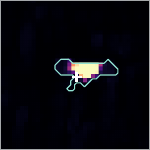}\end{minipage} & \begin{minipage}[c][2.20cm][c]{2.20cm}\centering\includegraphics[width=2.20cm,height=2.20cm,keepaspectratio]{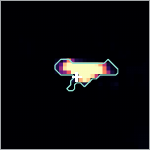}\end{minipage}\\[2pt]
\end{tabular}\par
\vspace{2pt}\includegraphics[width=3.4cm,height=.13cm]{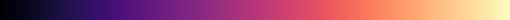}\\[-1pt]
\makebox[3.4cm]{0\hfill 0.5\hfill 1}}
\caption{Visualization of point-supported feature relations ($WA$) and detector predictions ($q$) with and without online affinity learning. Exploiting local feature similarity around annotated points (white crosses), online affinity learning adapts to varied shapes and recovers complete targets within ground-truth contours (cyan).}
\label{fig:affinity}
\end{figure*}

\paragraph{Implementation details} All main experiments use MSDA-Net \citep{zhao2025msda} with $256\times256$ crops, batch size 16, and 400 epochs on an NVIDIA RTX 3090 GPU. All OABL detectors are initialized from scratch without pretrained weights. AdamW starts at a learning rate of $10^{-3}$ with weight decay $10^{-4}$ and cosine scheduling. We fix $(m,\gamma)=(0.7,2)$, $(\sigma_s,\sigma_I)=(5,0.05)$, $(R,\tau)=(20,0.05)$, and $(\lambda_p,\lambda_a,\lambda_b)=(2,1,2)$ for all datasets. We run seeds 42, 2025, and 3407 and report the mean and sample standard deviation.

\paragraph{Compared methods} PAL and OABL use MSDA-Net, identical splits, and a 400-epoch budget. PAL retains its progressive schedule \citep{yu2025}. MCLC masks are generated by Monte Carlo linear clustering and DenseCRF before training MSDA-Net \citep{li2023mclc}. LESPS uses DNANet with joint SIRST3 training, and Diffuse to Detect uses ALCLNet \citep{ying2023,liu2026}. Full-mask MSDA-Net supplies the fully supervised reference under shared evaluation.

\paragraph{Evaluation} We extract eight-connected components and match predictions to ground truth one-to-one at centroid distances below three pixels. Detection probability $P_d$ is the fraction of matched targets. False-alarm area $F_a$ is the total area of unmatched predicted components divided by image area. We report results at threshold 0.5 and interpolate $P_d$ at $F_a=2\times10^{-5}$ between adjacent measured operating points bracketing that budget. Each threshold is applied to the entire dataset. Dataset-aggregated IoU is denoted mIoU, and mean per-image IoU is denoted nIoU. Target-pixel recall measures foreground coverage. Full-background error $F_{\mathrm{bg}}$ counts every predicted foreground pixel on true background, including excess pixels attached to matched components, divided by total image area. Area ratio is predicted foreground area divided by true foreground area. Unless specified otherwise, results use the epoch-400 checkpoint. The numerical stabilizers are $\epsilon=10^{-6}$ and $\epsilon_F=10^{-12}$.

\subsection{Target detection in challenging scenes}

\paragraph{Detection under strict false-alarm constraints} Table~\ref{tab:main} reports the detection capability learned from a single point per target. At $F_a=2\times10^{-5}$, OABL reaches $P_d$ of 97.28\%, 98.48\%, 94.42\%, and 93.38\% on SIRST3, NUDT-SIRST, NUAA-SIRST, and IRSTD-1K. These are gains of 1.06, 1.14, 7.82, and 6.07 percentage points over matched PAL training. At threshold 0.5, OABL also achieves the highest $P_d$ and lowest $F_a$ on SIRST3 and NUDT-SIRST while matching Diffuse to Detect in fixed-budget detection on NUAA-SIRST.

Fig.~\ref{fig:detection} tracks detection probability across false-alarm budgets. OABL maintains strong target retrieval over a broad part of the evaluated $P_d$--$F_a$ curves. These operating curves show that faint targets can be recovered directly from point annotations while controlling how much clutter is accepted as foreground.

\paragraph{Faint targets and complete scenes} Individual target matches do not capture whether a scene has missing detections. Compared with PAL, OABL raises complete-image detection from 85.83\% to 93.46\% on NUAA-SIRST and from 85.74\% to 91.38\% on IRSTD-1K. Table~\ref{tab:hard} examines low-contrast, low-SNR, and tiny targets at the same dataset-level operating point. On NUDT-SIRST, detection exceeds 96.6\% for both weak-signal subsets and reaches 98.58\% for targets occupying 1--10 pixels.

Target contrast is its mean intensity minus the median intensity of a surrounding background ring. SNR divides this contrast by the ring standard deviation plus $1/255$. The ring extends two to seven pixels beyond the target and excludes other targets. The lowest third of the complete target population defines each weak-signal subset. Fig.~\ref{fig:qualitative} presents matched examples with weak contrast, neighboring targets, and elongated responses. In the first scene, OABL retrieves a compact target beneath a brighter structure missed by PAL and MCLC. In the second, it resolves all three nearby targets. In the third, its response covers part of an elongated target missed by both baselines without false alarms.

\begin{table}[t]\centering\footnotesize
\caption{Detection probability ($P_d$, $\%$) of OABL on challenging subsets (low contrast, low SNR, and 1--10 pixel targets) at $F_a=2\times10^{-5}$, where $N$ is the number of targets.}\label{tab:hard}
\setlength{\tabcolsep}{2.5pt}\renewcommand{\arraystretch}{1.10}\begin{tabular*}{\columnwidth}{@{\extracolsep{\fill}}llrr@{}}\toprule
Dataset & Subset & $N$ & $P_d$ (\%)\\\midrule
NUDT-SIRST & Low contrast &315&97.46\\
&Low SNR&315&96.61\\
&1--10 pixels&259&98.58\\\midrule
IRSTD-1K&Low contrast&99&85.52\\
&Low SNR&99&83.17\\
&1--10 pixels&37&77.48\\\bottomrule
\end{tabular*}\end{table}

\subsection{Learning target structure from point-anchored features}\label{sec:affinity-analysis}

Table~\ref{tab:affinity} compares OABL with the same point and background objectives but without the affinity term ($\text{w/o affinity}$). On NUDT-SIRST, the affinity objective raises target-pixel recall from 79.36\% to 82.89\% and detection probability from 98.06\% to 98.24\%. The mean learned affinity within true targets rises from 0.709 to 0.988. Fig.~\ref{fig:affinity} makes this feature-level change visible under identical point annotations. With affinity learning, locally supported relations become stronger around the annotated targets. The paired prediction maps show more complete activation of a compact weak target and broader coverage along two elongated targets from sparse point supervision.

\begin{table}[t]\centering\footnotesize
\caption{Ablation study of online foreground affinity learning on NUDT-SIRST ($\%$). Adding the affinity objective improves target-pixel recall and detection rate.}\label{tab:affinity}
\setlength{\tabcolsep}{2.5pt}\renewcommand{\arraystretch}{1.10}\begin{tabular*}{\columnwidth}{@{\extracolsep{\fill}}lrrrr@{}}\toprule
Objective & Pixel recall & $P_d$ & mIoU & nIoU\\\midrule
w/o affinity&79.36&98.06&62.95&67.20\\
w/ affinity (OABL)&82.89&98.24&62.46&66.56\\\bottomrule
\end{tabular*}\end{table}

\paragraph{Feature alignment versus output matching} A learned affinity relation can guide shared features or serve as an explicit target for the output map. We evaluate these two designs by adding prediction-to-affinity matching $C$ in Eq.~\eqref{eq:consistency} to the OABL objective. Table~\ref{tab:matching} shows higher nIoU for direct feature alignment in OABL ($\mathcal L$) across all four settings, with gains of 1.03, 3.67, 2.10, and 2.35 percentage points on SIRST3, NUDT-SIRST, NUAA-SIRST, and IRSTD-1K. OABL uses online affinity to shape the shared representation and learns pixel probabilities from the joint point and background losses.

\begin{equation}\label{eq:consistency}
C=\frac{1}{B|\Omega|}\sum_{b,u}V_b\bigl(q_b(u)-\sg(A_b(u))\bigr)^2,
\end{equation}
where $V_b=\ind[|\mathcal P_b|>0]$. Replacing $A_b$ with fixed support $W_b$ in Eq.~\eqref{eq:consistency} gives the fixed-prior matching variant $C_W$.

\begin{table}[t]\centering\footnotesize
\caption{Comparison between feature-level affinity alignment ($\mathcal L$) and explicit output-level matching ($\mathcal L+C$) across four benchmarks ($\%$). Guiding shared features yields superior region quality (nIoU).}\label{tab:matching}
\setlength{\tabcolsep}{2.5pt}\renewcommand{\arraystretch}{1.10}\begin{tabular*}{\columnwidth}{@{\extracolsep{\fill}}llrrrr@{}}\toprule
Dataset & Objective & \shortstack{$P_d$ at\\fixed $F_a$} & $P_d$ & mIoU & nIoU\\\midrule
SIRST3&$\mathcal L+C$&97.12&97.05&64.50&65.93\\
&$\mathcal L$ (OABL)&97.28&97.08&63.44&66.96\\\midrule
NUDT-SIRST&$\mathcal L+C$&98.45&98.10&54.59&62.89\\
&$\mathcal L$ (OABL)&98.48&98.24&62.46&66.56\\\midrule
NUAA-SIRST&$\mathcal L+C$&93.43&92.90&55.09&61.00\\
&$\mathcal L$ (OABL)&94.42&93.54&56.47&63.10\\\midrule
IRSTD-1K&$\mathcal L+C$&94.39&94.28&57.57&55.62\\
&$\mathcal L$ (OABL)&93.38&93.15&57.78&57.97\\\bottomrule
\end{tabular*}\end{table}

\begin{table}[t]\centering\footnotesize
\caption{Ablation study on how local image support ($W$) enters training on SIRST3. Using $W$ to weight feature alignment ($\mathcal L_{\mathrm{aff}}$) outperforms fitting $W$ or $A$ as an output target ($C_W, C$), achieving the highest fixed-budget detection rate and lowest background error ($F_{\mathrm{bg}}$, $\times 10^{-5}$).}\label{tab:support}
\setlength{\tabcolsep}{2.5pt}\renewcommand{\arraystretch}{1.10}\begin{tabular*}{\columnwidth}{@{\extracolsep{\fill}}lrrrrr@{}}\toprule
Added term & \shortstack{$P_d$ at\\fixed $F_a$} & $F_{\mathrm{bg}}$ & \shortstack{Pixel\\recall} & mIoU & nIoU\\\midrule
$C$ &97.14&10.46&77.55&63.24&66.51\\
$C_W$, $\sigma_s=5$&96.86&10.19&78.48&64.30&67.01\\
$C_W$, $\sigma_s=10$&97.19&9.74&77.04&63.64&67.36\\
$\mathcal L_{\mathrm{aff}}+C$&97.12&13.02&82.66&64.50&65.93\\
$\mathcal L_{\mathrm{aff}}$ (OABL)&97.28&9.69&76.74&63.44&66.96\\\bottomrule
\end{tabular*}\end{table}

\begin{table*}[t]\centering\footnotesize
\caption{Comparison of background supervision objectives on NUDT-SIRST ($P_d$ and mIoU in $\%$, $F_a$ and $F_{\mathrm{bg}}$ in $10^{-5}$). By addressing the long-tailed distribution of residual clutter, active-set averaging (OABL) achieves high detection accuracy with compact predictions and low false alarms under both threshold 0.5 and the $F_a\le2\times10^{-5}$ budget.}\label{tab:background}
\setlength{\tabcolsep}{5pt}\begin{tabular}{lrrrrrrr}\toprule
&\multicolumn{4}{c}{Default threshold 0.5}&\multicolumn{3}{c}{Low-false-alarm budget}\\
\cmidrule(lr){2-5}\cmidrule(lr){6-8}
Background objective&$P_d$&$F_a$&mIoU&Area ratio&$P_d$&mIoU&$F_{\mathrm{bg}}$\\\midrule
Active mean (OABL)&98.24&0.65&62.46&1.16&98.48&62.28&25.67\\
Candidate mean, $\lambda=2$&92.77&149.51&10.66&9.29&---&---&---\\
Candidate mean, $\lambda=36$&98.20&11.35&30.40&3.27&97.71&44.40&71.68\\
Candidate BCE, $\lambda=2$&92.87&65.74&17.80&5.71&97.11&44.26&66.68\\\bottomrule
\end{tabular}\par\vspace{3pt}\footnotesize ---: the candidate-mean variant with $\lambda=2$ does not reach the $F_a\le2\times10^{-5}$ budget.
\end{table*}

\begin{figure*}[!t]
\centering
{\footnotesize\swatch{blue}{solid}\ NUDT-SIRST\qquad\swatch{copper}{solid}\ IRSTD-1K\par}\vspace{2pt}
\begin{minipage}{.94\textwidth}\centering
\begin{subfigure}{.32\linewidth}\centering
\includegraphics[width=\linewidth]{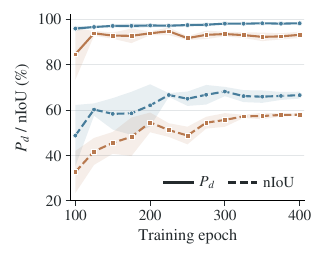}
\caption{Target learning}\label{fig:dynamics-target}
\end{subfigure}\hfill
\begin{subfigure}{.32\linewidth}\centering
\includegraphics[width=\linewidth]{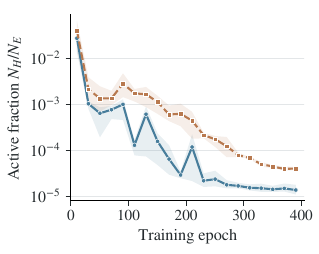}
\caption{Active background fraction}\label{fig:dynamics-fraction}
\end{subfigure}\hfill
\begin{subfigure}{.32\linewidth}\centering
\includegraphics[width=\linewidth]{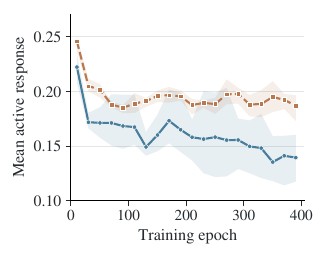}
\caption{Residual background response}\label{fig:dynamics-response}
\end{subfigure}
\end{minipage}
\caption{Single-stage training dynamics on NUDT-SIRST and IRSTD-1K. Target detection ($P_d$) and region quality (nIoU) improve steadily during training (a), while the active background fraction shrinks by over three orders of magnitude (b) and active normalization sustains supervision on residual clutter (c).}
\label{fig:dynamics}
\end{figure*}

Table~\ref{tab:support} further examines how spatial and intensity information enters training on SIRST3 under matched point and background losses. Fitting the detector output directly to the fixed support map $W$ at $\sigma_s=5$ or $\sigma_s=10$ ($C_W$) or to detached affinity $A$ ($C$) treats uncertain local cues as explicit output targets. Adding $C$ on top of $\mathcal L_{\mathrm{aff}}$ raises pixel recall to 82.66\% but increases full-background error $F_{\mathrm{bg}}$ to $13.02\times10^{-5}$. Using $W$ solely to weight feature alignment ($\mathcal L_{\mathrm{aff}}$) achieves the highest fixed-budget detection rate (97.28\%) and the lowest background error ($9.69\times10^{-5}$).

\subsection{Active background learning for residual clutter}

We next examine what happens when most eligible background pixels become easy to suppress and only a few target-like responses remain. With the point and affinity objectives fixed, we compare OABL's active-set mean with candidate-wide probability averaging and binary cross-entropy (BCE) over $E_b$.

\paragraph{Suppressing target-like background responses} Table~\ref{tab:background} shows that active averaging yields compact predictions while retaining high target detection. On NUDT-SIRST, OABL reaches $P_d=98.24\%$, $F_a=0.65\times10^{-5}$, and an area ratio of 1.16. Candidate-wide probability averaging with the same weight ($\lambda=2$) dilutes gradients on hard clutter and expands the area ratio to 9.29 ($F_a=149.51\times10^{-5}$). Even after increasing its weight from 2 to 36, detection reaches a similar 98.20\% but $F_a$ remains $11.35\times10^{-5}$ with an area ratio of 3.27. The active mean therefore focuses negative learning where residual high responses still require correction.

We also compare these background objectives at the same false-alarm budget in Table~\ref{tab:background}, which reports the best measured operating point satisfying $F_a\le 2\times10^{-5}$ for each seed. Active averaging achieves 98.48\% detection and 62.28\% mIoU with $F_{\mathrm{bg}}=25.67\times10^{-5}$, whereas the candidate-wide variants reach 97.71\% and 97.11\% detection with mIoU near 44\% and nearly triple the background error. These operating-point results confirm that active normalization improves region quality even when decision thresholds are adjusted.

\paragraph{Residual clutter dynamics and active normalization} The active fraction $N_H/N_E$ in Fig.~\ref{fig:dynamics}(b) falls from 2.77\% to 0.00137\% on NUDT-SIRST and from 4.05\% to 0.00404\% on IRSTD-1K between the first and final recorded 20-epoch intervals. Meanwhile, the remaining selected pixels retain mean foreground probabilities of 0.139 and 0.187, respectively (Fig.~\ref{fig:dynamics}(c)). Thus the number of unresolved responses contracts sharply while the active-set mean continues to measure their remaining strength.

Selecting hard responses changes the loss numerator, whereas averaging over the selected count changes its denominator. Table~\ref{tab:normalization} isolates these two operations under a common point and affinity objective that includes the output-matching term $C$. At $\lambda=2$, replacing the active denominator $N_H$ with all eligible pixels $N_E$ ($B_{H/E}$) reduces mIoU from 54.59\% to 2.58\% and expands the predicted area ratio from 1.39 to 38.63, closely matching unselective candidate averaging ($B_E$). Even when $\lambda$ is increased to 8 and 32 for $B_{H/E}$ and $B_E$, false-alarm area $F_a$ remains above $120\times10^{-5}$ and the predicted area exceeds $23\times$ the true target area. Normalizing by the active count $N_H$ is therefore the decisive mechanism in Table~\ref{tab:normalization}.

\begin{equation}\label{eq:normalization}
B_H=\frac{\sum Hq}{N_H+\epsilon},\quad B_{H/E}=\frac{\sum Hq}{N_E+\epsilon},\quad B_E=\frac{\sum Eq}{N_E+\epsilon}.
\end{equation}
Sums run over the batch and pixel grid. $E$ is the candidate indicator and $N_E=\sum_b|E_b|$.

\begin{table}[!htbp]\centering\footnotesize
\caption{Ablation study isolating hard-negative selection ($H$) and active-count normalization ($N_H$) on NUDT-SIRST ($P_d$ and mIoU in $\%$, $F_a$ and $F_{\mathrm{bg}}$ in $10^{-5}$). Normalizing by the active count ($B_H$) instead of the full candidate set ($B_{H/E}, B_E$) sustains effective supervision on residual false alarms.}\label{tab:normalization}
\setlength{\tabcolsep}{2.5pt}\renewcommand{\arraystretch}{1.10}\begin{tabular*}{\columnwidth}{@{\extracolsep{\fill}}llrrrrr@{}}\toprule
Reduction&$\lambda$&$P_d$&$F_a$&mIoU&$F_{\mathrm{bg}}$&\shortstack{Area\\ratio}\\\midrule
$B_H$&2&98.10&1.51&54.59&36.17&1.39\\\midrule
$B_{H/E}$&2&83.81&420.43&2.58&2448.63&38.63\\
&8&87.05&204.27&3.21&1947.13&30.92\\
&32&89.52&120.50&4.25&1456.20&23.37\\\midrule
$B_E$&2&83.07&481.36&2.54&2488.28&39.24\\
&8&87.20&212.54&3.19&1962.25&31.15\\
&32&87.16&142.92&4.21&1470.16&23.59\\\bottomrule
\end{tabular*}\end{table}

\begin{figure}[!htbp]
\centering
{\footnotesize\swatch{oablred}{solid}\ $2B_H$\quad\swatch{blue}{dashed}\ $32B_{H/E}$\quad\swatch{methodgray}{dash dot}\ $32B_E$\par}\vspace{2pt}
\begin{subfigure}{.485\columnwidth}\centering
\includegraphics[width=\linewidth]{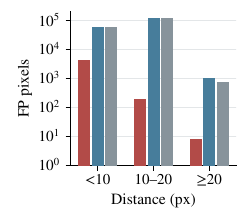}
\caption{False-positive pixels}
\end{subfigure}\hfill
\begin{subfigure}{.485\columnwidth}\centering
\includegraphics[width=\linewidth]{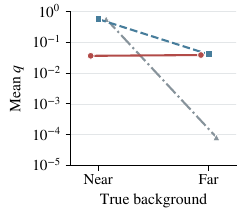}
\caption{Background response}
\end{subfigure}
\caption{Comparison of false-positive pixel counts (a) and mean background responses (b) across distance bands on NUDT-SIRST. Active-set normalization ($2B_H$) sharply reduces false alarms and suppresses residual background responses both near ($<20$ px) and far ($\ge20$ px) from target points compared with candidate-wide averaging ($32B_{H/E}$ and $32B_E$).}
\label{fig:spatialerrors}
\end{figure}

\paragraph{Where residual errors appear} Fig.~\ref{fig:spatialerrors} shows the number, location, and strength of false foreground responses under the three matched background reductions. Across the diagnostic images, $2B_H$ produces 4,445 false-positive pixels, compared with 181,415 for $32B_{H/E}$ and 181,610 for $32B_E$. The all-candidate mean ($32B_E$) strongly suppresses distant background while retaining high responses near the excluded target neighborhoods, where near-point background probability averages 0.584 and 65.45\% of false-positive pixels lie 10--20 pixels from a point. Under $2B_H$, these values fall to 0.0369 and 4.34\%. Active supervision selected outside radius $R$ also regularizes nearby predictions through the shared network parameters.

\subsection{Joint training and computational cost}

The same OABL objective trains the detector for 400 epochs. Fig.~\ref{fig:dynamics}(a) follows target detection and regional agreement from epoch 100 onward. On NUDT-SIRST, $P_d$ rises from 95.91\% to 98.24\% and nIoU from 48.75\% to 66.56\%. On IRSTD-1K, the corresponding values increase from 84.62\% to 93.15\% and from 32.70\% to 57.97\%. These gains develop alongside the contraction of the active background population in Fig.~\ref{fig:dynamics}(b,c) within the same single-stage optimization process.

\begin{table}[!htbp]\centering\footnotesize
\caption{Comparison of training complexity with PAL and computational cost of OABL over 400 epochs. OABL simplifies training into a single stage with a low memory footprint.}\label{tab:cost}
\renewcommand{\arraystretch}{1.06}
\begin{tabular*}{\columnwidth}{@{\extracolsep{\fill}}lrr@{}}\toprule
\multicolumn{3}{@{}l}{\textit{(a) Training organization}}\\[2pt]
Property & PAL & OABL\\\midrule
Training stages & 3 & 1\\
Extra full-pool passes & 111 & 0\\
Persistent pseudo-masks & Yes & No\\
Inference parameters (M) & 4.81 & 4.81\\\midrule
\multicolumn{3}{@{}l}{\textit{(b) OABL resource measurements}}\\[2pt]
Dataset & Time (h) & Memory (GiB)\\\midrule
SIRST3 & $9.20\pm0.01$ & 3.35\\
NUDT-SIRST & $3.79\pm0.13$ & 2.94\\
NUAA-SIRST & $1.24\pm0.05$ & 3.35\\
IRSTD-1K & $4.33\pm0.03$ & 2.82\\\bottomrule
\end{tabular*}
\end{table}

OABL obtains both learning signals from the current forward pass and updates the detector once per mini-batch. Table~\ref{tab:cost} compares this procedure with matched PAL training. PAL uses three stages and 111 additional full-pool inference passes to generate and refresh labels, whereas OABL uses a single stage and zero extra passes while retaining the same 4.81M-parameter detector. Panel (b) reports CUDA-event training time (1.24--9.20 hours, excluding evaluation and data-loading waits) and peak reserved GPU memory (2.82--3.35 GiB) across the four benchmarks. Recomputing point affinities and active background masks online within each mini-batch avoids storing or reloading intermediate label maps across training stages.

\FloatBarrier
\section{Conclusion}

We presented OABL, a single-stage training framework for infrared small-target detection from one point per target. The core idea is to use local spatial and intensity cues as weights on point-anchored feature alignment while relying on current predictions to select high-response background clutter outside uncertain target neighborhoods. Online foreground affinity learning builds target-related representations within the shared backbone, and active-set normalization sustains negative supervision on residual false alarms as easy background pixels vanish during training. Both mechanisms update the same detector in each forward pass without intermediate pseudo-mask generation.

Across NUDT-SIRST, NUAA-SIRST, IRSTD-1K, and the combined SIRST3 benchmark, OABL achieves strong target retrieval across low false-alarm operating regimes while keeping the original inference network unchanged. Feature visualizations, objective comparisons, and spatial error diagnostics verify how local image support and active background normalization complement each other during optimization. Together, these findings show that assigning local image evidence to representation learning and hard background feedback to prediction learning provides a direct, effective route from single-point annotations to pixel-level infrared small-target detection.


\begin{thebibliography}{99}
\bibitem[Ahn et~al.(2019)]{ahn2019}
Ahn, J., Cho, S., Kwak, S., 2019. Weakly supervised learning of instance segmentation with inter-pixel relations, in: Proceedings of the IEEE/CVF Conference on Computer Vision and Pattern Recognition (CVPR), pp. 2209--2218. doi:10.1109/CVPR.2019.00231.

\bibitem[Ahn and Kwak(2018)]{ahn2018}
Ahn, J., Kwak, S., 2018. Learning pixel-level semantic affinity with image-level supervision for weakly supervised semantic segmentation, in: Proceedings of the IEEE Conference on Computer Vision and Pattern Recognition (CVPR), pp. 4981--4990. doi:10.1109/CVPR.2018.00523.

\bibitem[Bearman et~al.(2016)]{bearman2016}
Bearman, A., Russakovsky, O., Ferrari, V., Fei-Fei, L., 2016. What's the point: Semantic segmentation with point supervision, in: Computer Vision -- ECCV 2016, Springer International Publishing. pp. 549--565. doi:10.1007/978-3-319-46478-7\_34.

\bibitem[Cao et~al.(2020)]{cao2020}
Cao, Y., Chen, K., Loy, C.C., Lin, D., 2020. Prime sample attention in object detection. IEEE/CVF Conference on Computer Vision and Pattern Recognition (CVPR), 11583--11591.

\bibitem[Dai et~al.(2021)]{dai2021}
Dai, Y., Wu, Y., Zhou, F., Barnard, K., 2021. Asymmetric contextual modulation for infrared small target detection, in: Proceedings of the IEEE/CVF Winter Conference on Applications of Computer Vision (WACV), pp. 950--959. doi:10.1109/WACV48630.2021.00099.

\bibitem[Gao et~al.(2026)]{gao2026}
Gao, W., Niu, W., Tang, J., Yang, M., Zhang, J., Peng, X., 2026. Point-to-mask: From arbitrary point annotations to mask-level infrared small target detection. doi:10.48550/arXiv.2603.16257, arXiv:2603.16257.

\bibitem[Kou et~al.(2024)]{kou2024}
Kou, R., Wang, C., Fu, Q., Li, Z., Luo, Y., Li, B., Li, W., Peng, Z., 2024. MCGC: A multiscale chain growth clustering algorithm for generating infrared small target mask under single-point supervision. IEEE Transactions on Geoscience and Remote Sensing 62, 1--12. doi:10.1109/TGRS.2024.3390756.

\bibitem[Li et~al.(2019)]{li2019ghm}
Li, B., Liu, Y., Wang, X., 2019. Gradient harmonized single-stage detector. AAAI Conference on Artificial Intelligence 33, 8577--8584. doi:10.1609/aaai.v33i01.33018577.

\bibitem[Li et~al.(2023a)]{li2023mclc}
Li, B., Wang, Y., Wang, L., Zhang, F., Liu, T., Lin, Z., An, W., Guo, Y., 2023a. Monte carlo linear clustering with single-point supervision is enough for infrared small target detection, in: Proceedings of the IEEE/CVF International Conference on Computer Vision (ICCV), pp. 1009--1019. doi:10.1109/ICCV51070.2023.00099.

\bibitem[Li et~al.(2023b)]{li2023dnanet}
Li, B., Xiao, C., Wang, L., Wang, Y., Lin, Z., Li, M., An, W., Guo, Y., 2023b. Dense nested attention network for infrared small target detection. IEEE Transactions on Image Processing 32, 1745--1758. doi:10.1109/TIP.2022.3199107.

\bibitem[Li et~al.(2024)]{li2024levelset}
Li, H., Yang, J., Xu, Y., Wang, R., 2024. A level set annotation framework with single-point supervision for infrared small target detection. IEEE Signal Processing Letters 31, 451--455. doi:10.1109/LSP.2024.3356411.

\bibitem[Li et~al.(2023c)]{li2023point2mask}
Li, W., Yuan, Y., Wang, S., Zhu, J., Li, J., Liu, J., Zhang, L., 2023c. Point2mask: Point-supervised panoptic segmentation via optimal transport, in: Proceedings of the IEEE/CVF International Conference on Computer Vision (ICCV), pp. 572--581. doi:10.1109/ICCV51070.2023.00059.

\bibitem[Li and Zhang(2026)]{li2026teacher}
Li, Y., Zhang, J., 2026. Teacher-guided progressive learning for infrared small target detection with single-point supervision. Infrared Physics \& Technology 157, 106667. doi:10.1016/j.infrared.2026.106667.

\bibitem[Lin et~al.(2017)]{lin2017}
Lin, T.Y., Goyal, P., Girshick, R., He, K., Dollár, P., 2017. Focal loss for dense object detection, in: Proceedings of the IEEE International Conference on Computer Vision (ICCV), pp. 2980--2988.

\bibitem[Liu et~al.(2026)]{liu2026}
Liu, Z., Yao, Y., Qian, P., Chen, Z., Liu, R., 2026. Diffuse to detect: Bi-level sample rebalancing with pseudo-label diffusion for point-supervised infrared small-target detection, in: Proceedings of the 43rd International Conference on Machine Learning, pp. 79394-- 79413.

\bibitem[Ni et~al.(2026)]{ni2026}
Ni, R., Li, B., Chen, J., Li, Y., Ren, F., Wang, Y., Yuan, H., He, W., An, W., 2026. Rethinking IRSTD: Single-point supervision guided encoder-only framework is enough for infrared small target detection. doi:10.48550/arXiv.2604.05363, arXiv:2604.05363.

\bibitem[Peng et~al.(2025)]{peng2025}
Peng, J., Zhuang, S., Qi, M., Li, K., Zhang, K., Yang, Y., 2025. Single-point-supervised infrared small target detection with local affinity feature refinement, in: Seventeenth International Conference on Digital Image Processing (ICDIP 2025), SPIE. p. 137091B. doi:10.1117/12.3073106.

\bibitem[Ru et~al.(2022)]{ru2022}
Ru, L., Zhan, Y., Yu, B., Du, B., 2022. Learning affinity from attention: End-to-end weakly-supervised semantic segmentation with transformers, in: Proceedings of the IEEE/CVF Conference on Computer Vision and Pattern Recognition (CVPR), pp. 16846--16855.

\bibitem[Shrivastava et~al.(2016)]{shrivastava2016}
Shrivastava, A., Gupta, A., Girshick, R., 2016. Training region-based object detectors with online hard example mining, in: Proceedings of the IEEE Conference on Computer Vision and Pattern Recognition (CVPR), pp. 761--769.

\bibitem[Tang et~al.(2018)]{tang2018}
Tang, M., Perazzi, F., Djelouah, A., Ben Ayed, I., Schroers, C., Boykov, Y., 2018. On regularized losses for weakly-supervised CNN segmentation. European Conference on Computer Vision (ECCV), 507--522.

\bibitem[Wang et~al.(2022)]{wang2022}
Wang, Y., Wang, H., Shen, Y., Fei, J., Li, W., Jin, G., Wu, L., Zhao, R., Le, X., 2022. Semi-supervised semantic segmentation using unreliable pseudo-labels. IEEE/CVF Conference on Computer Vision and Pattern Recognition (CVPR), 4248--4257.

\bibitem[Wu et~al.(2023)]{wu2023}
Wu, X., Hong, D., Chanussot, J., 2023. UIU-Net: U-Net in U-Net for infrared small object detection. IEEE Transactions on Image Processing 32, 364--376. doi:10.1109/TIP.2022.3228497.

\bibitem[Wu et~al.(2016)]{wu2016}
Wu, Z., Shen, C., van den Hengel, A., 2016. High-performance semantic segmentation using very deep fully convolutional networks. arXiv:1604.04339.

\bibitem[Yang et~al.(2024)]{yang2024}
Yang, D., Zhang, H., Li, Y., Jiang, Z., 2024. Label evolution based on local contrast measure for single-point-supervised infrared small-target detection. IEEE Transactions on Geoscience and Remote Sensing 62, 1--12. doi:10.1109/TGRS.2024.3472455.

\bibitem[Ying et~al.(2023)]{ying2023}
Ying, X., Liu, L., Wang, Y., Li, R., Chen, N., Lin, Z., Sheng, W., Zhou, S., 2023. Mapping degeneration meets label evolution: Learning infrared small target detection with single point supervision, in: Proceedings of the IEEE/CVF Conference on Computer Vision and Pattern Recognition (CVPR), pp. 15528--15538. doi:10.1109/CVPR52729.2023.01490.

\bibitem[Yu et~al.(2025)]{yu2025}
Yu, C., Zhao, J., Liu, Y., Zhao, S., Dai, Y., Yue, X., 2025. From easy to hard: Progressive active learning framework for infrared small target detection with single point supervision, in: Proceedings of the IEEE/CVF International Conference on Computer Vision (ICCV), pp. 2588--2598. doi:10.1109/ICCV51701.2025.00249.

\bibitem[Zhang et~al.(2022)]{zhang2022}
Zhang, M., Zhang, R., Yang, Y., Bai, H., Zhang, J., Guo, J., 2022. ISNet: Shape matters for infrared small target detection, in: Proceedings of the IEEE/CVF Conference on Computer Vision and Pattern Recognition (CVPR), pp. 877--886. doi:10.1109/CVPR52688.2022.00095.

\bibitem[Zhao et~al.(2025a)]{zhao2025fest}
Zhao, J., Shi, Z., Yu, C., Liu, Y., Dai, Y., 2025a. Towards robust infrared small target detection: A feature-enhanced and sensitivity-tunable framework. Knowledge-Based Systems 330, 114519. doi:10.1016/j.knosys.2025.114519.

\bibitem[Zhao et~al.(2025b)]{zhao2025msda}
Zhao, J., Shi, Z., Yu, C., Liu, Y., Ying, X., Dai, Y., 2025b. Multiscale direction-aware network for infrared small target detection. IEEE Transactions on Geoscience and Remote Sensing 63, 1--18. doi:10.1109/TGRS.2025.3612039.

\end{thebibliography}
\end{document}